\documentclass[letterpaper, 10 pt, conference]{ieeeconf} 

\IEEEoverridecommandlockouts                             
\usepackage{hyperref}
\usepackage{verbatim}
\usepackage{stfloats}
\usepackage{makecell}
\usepackage{array}
\usepackage{ulem}
\usepackage{bbding}
\usepackage{microtype}
\usepackage{subfigure}
\usepackage{booktabs}

\usepackage{graphics}
\usepackage{float}
\usepackage{epsfig} 
\usepackage{mathptmx}
\usepackage{times} 
\usepackage{amsmath}
\usepackage{amssymb} 
\usepackage{pifont}
\usepackage{cite}

\usepackage[table]{xcolor}
\usepackage{ulem}
\usepackage[utf8]{inputenc}
\usepackage{algorithm}
\usepackage{algpseudocode}

\author{Weijie Zhou$^{1,2}$, Manli Tao$^{2}$, Chaoyang Zhao$^{2,3*}$, Honghui Dong$^{1}$, Ming Tang$^{2}$,    Jinqiao Wang$^{2,3}$
\thanks{$^{*}$Corresponding Author (Email: chaoyang.zhao@nlpr.ia.ac.cn)}
\thanks{$^{1}$School of Traffic and Transportation, Beijing Jiaotong University}
\thanks{$^{2}$Foundation Model Research Center, Institute of Automation, Chinese Academy of Sciences}
\thanks{$^{3}$objecteye.Inc}%
\thanks{Supplementary material, model and the code: \href{https://github.com/jetteezhou/LightPlanner}{\url{https://github.com/jetteezhou/LightPlanner}}}%
}

\title{\LARGE \bf
LightPlanner: Unleashing the Reasoning Capabilities of Lightweight Large Language Models in Task Planning}

\begin{document}

\maketitle
\thispagestyle{empty}
\pagestyle{empty}

\begin{abstract}
In recent years, lightweight large language models (LLMs) have garnered significant attention in the robotics field due to their low computational resource requirements and suitability for edge deployment. However, in task planning—particularly for complex tasks that involve dynamic semantic logic reasoning—lightweight LLMs have underperformed. To address this limitation, we propose a novel task planner, LightPlanner, which enhances the performance of lightweight LLMs in complex task planning by fully leveraging their reasoning capabilities. Unlike conventional planners that use fixed skill templates, LightPlanner controls robot actions via parameterized function calls, dynamically generating parameter values. This approach allows for fine-grained skill control and improves task planning success rates in complex scenarios. Furthermore, we introduce hierarchical deep reasoning. Before generating each action decision step, LightPlanner thoroughly considers three levels: action execution (feedback verification), semantic parsing (goal consistency verification), and parameter generation (parameter validity verification). This ensures the correctness of subsequent action controls. Additionally, we incorporate a memory module to store historical actions, thereby reducing context length and enhancing planning efficiency for long-term tasks. We train the LightPlanner-1.5B model on our LightPlan-40k dataset, which comprises 40,000 action controls across tasks with 2 to 13 action steps. Experiments demonstrate that our model achieves the highest task success rate despite having the smallest number of parameters. In tasks involving spatial semantic reasoning, the success rate exceeds that of ReAct by 14.9\%. Moreover, we demonstrate LightPlanner's potential to operate on edge devices.

\end{abstract}

\begin{figure}[ht] 
\vskip 0.2in
\begin{center}
\centerline{\includegraphics[width=1\columnwidth]{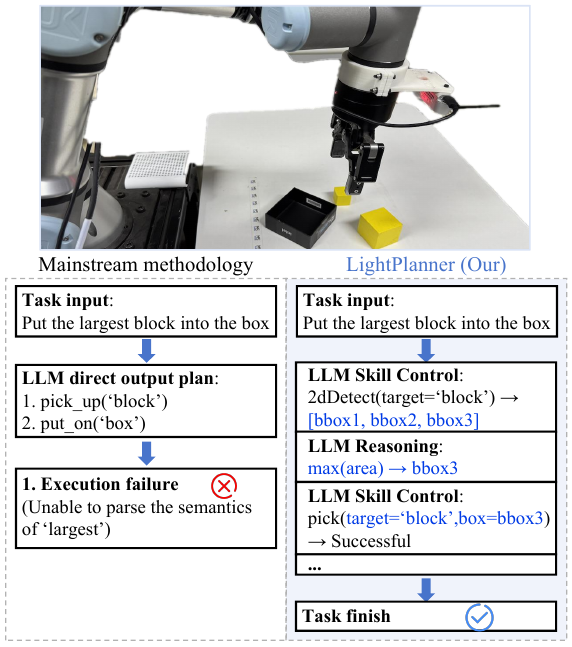}}
\caption{Mainstream LLM planning methods (left) rely on predefined templates that struggle to flexibly interpret semantics such as "largest." In contrast, LightPlanner (right) enables LLMs to actively generate dynamic parameters and skill controls. The key innovation lies in decomposing high-level instructions into parameterized skill chains (detect $\rightarrow$ reason $\rightarrow$ pick), where the LLM proactively parses the "largest" attribute through bounding box area calculations, ultimately enabling precise grasping.}
\label{image1}
\end{center}
\vskip -0.2in
\end{figure}

\section{Introduction}

Embodied task planning involves translating high-level language instructions into specific executable actions \cite{cap, cape, chatgpt-for-robotics, copal}. In robotics, the accuracy and efficiency of task planning directly impact real-world performance \cite{challenge2, replan2, replan1}. Recently, large language models (LLMs) have been increasingly integrated into robotic task planning due to their strong natural language understanding and generation capabilities \cite{llm-glm, llm-llama2, llmalpaca, llmreasoningsurvey}. However, high-performance LLMs typically have large parameter sizes and high computational demands, limiting their deployment and real-time application on resource-constrained edge devices \cite{challenge2, chanllenges}. To address this limitation, lightweight LLMs have been developed to significantly reduce computational resource consumption while maintaining high performance, making them more suitable for embedded systems like robots \cite{lightplan1, lightplan2, lightplan3, lightplan4}.

Despite their resource efficiency, lightweight LLMs underperform in complex task planning, especially for tasks requiring multi-step logical reasoning and dynamic parameter adjustments \cite{lightplan1, lightplan4, challenge2, chanllenges, corki-cloud-time}. For instance, in the task "grab the largest block from some blocks and place it in the box," the system must identify the target block, dynamically calculate semantic attributes (e.g., size relationships), and generate precise control parameters (e.g., coordinates of the largest block). The poor performance of lightweight LLMs primarily stems from limited reasoning depth and flexibility, making it difficult to handle complex semantic relationships and execution feedback effectively. This limitation reduces the success rate and robustness of task planning.

To address these challenges, we propose \textit{LightPlanner}, a novel task planner that enhances the performance of lightweight LLMs in complex task planning by fully leveraging their reasoning capabilities. Unlike traditional methods that rely on fixed skill templates \cite{chatgpt-for-robotics, fltrnn-replan, isrllm}, LightPlanner employs a parameterized function call mechanism to dynamically control the robot's actions, enabling finer skill control through dynamic parameter inference. This approach significantly improves task planning success rates in complex scenarios. Additionally, LightPlanner introduces a hierarchical deep reasoning mechanism. Before generating each action decision, the system engages in deep reasoning at three levels: action execution (feedback verification), semantic parsing (goal consistency verification), and parameter generation (parameter validity verification). This ensures the correctness and rationality of each action step, enhancing task planning accuracy and the system's ability to handle anomalies. To further improve long-term task planning efficiency, LightPlanner incorporates a memory module to store historical action records, reducing the need for extensive context. This design enables LightPlanner to utilize existing information more efficiently, achieving continuous and stable task execution in multi-step complex tasks.

To validate our method, we constructed the LightPlan-40k dataset, comprising 40,000 action controls across tasks with 2 to 13 action steps, and trained the \textit{LightPlanner-1.5B} model on it. Experimental results demonstrate that, in real-world environments, the LightPlanner model achieves the highest task success rate despite having the smallest number of parameters. In tasks requiring spatial semantic reasoning, its success rate exceeds that of ReAct by 14.9\%. Additionally, we demonstrate LightPlanner's potential for operation on edge devices, highlighting its applicability in resource-constrained environments.

In summary, our main contributions are:
\begin{enumerate}
    \item We propose \textit{LightPlanner}, a task planner based on lightweight LLMs, capable of achieving efficient and high success rate planning in complex task scenarios.
    \item We introduce dynamic parameterized skill control and a hierarchical deep reasoning mechanism, significantly improving the accuracy and robustness of task planning.
    \item We embed a memory module to store historical action records, reducing the need for extensive context and optimizing planning efficiency for long-term tasks.
    \item We construct the \textit{LightPlan-40k} dataset to train the \textit{LightPlanner-1.5B} model. Experiments in real environments validate the superior performance of the LightPlanner model in task success rates, particularly in tasks involving spatial semantic reasoning. Furthermore, we demonstrate the potential of LightPlanner to operate on edge devices, highlighting its application prospects in resource-constrained environments.
\end{enumerate}

\section{Related Work}

\subsection{LLMs for Embodied Task Planning}

Early research on using LLMs for robotic task planning focused on leveraging high-performance models like ChatGPT \cite{gpt3, gpt4} for instruction following and context understanding, enabling the generation of task plans based on examples \cite{cap, cape, fltrnn-replan, innermonologue-planner, copal, palme}. For instance, SayCan \cite{saycan} utilized PALM \cite{palm} as a high-level decision-maker for robots but relied on affordance functions to filter out infeasible actions. ChatGPT for Robotics \cite{chatgpt-for-robotics} explored task planning across various robotic tasks. Code as Policies \cite{cap} introduced a method using GPT-3 to generate policy code, allowing robots to perform tasks such as desktop operations. However, high-performance LLMs generally have large parameter sizes and high computational demands. Moreover, API-based LLMs require internet connectivity, limiting their deployment on resource-constrained edge devices.

With advancements in LLMs, some studies have fine-tuned open-source models \cite{llm-glm, llm-llama2, qwen2, ll} with smaller parameter sizes for long-term task planning and action decision-making \cite{plan-mldt, react}. MLDT \cite{plan-mldt} leverages GPT-3.5 to generate multi-layer task planning data and fine-tunes LLMs with 6B, 7B, and 13B parameters for long-term planning. ReAct \cite{react} introduced a reasoning-action paradigm for executing long-term tasks and fine-tuned LLMs with 8B and 62B parameters. Although these approaches address some challenges of open-source LLMs in long-term task planning, their effectiveness in complex task planning, especially for tasks requiring multi-step logical reasoning and dynamic parameter adjustments, remains limited. This paper explores the potential of a lightweight 1.5B parameter LLM, capable of real-time operation on edge devices, for complex task planning.

\subsection{Task Planner with Reasoning or Replanning}

To improve task planning success rates, some studies have incorporated reasoning mechanisms into conventional planning processes \cite{innermonologue-planner, interactive-planning, llmreasoningsurvey}. Inner Monologue \cite{innermonologue-planner} uses LLMs and a self-questioning mechanism to enhance multi-step planning success rates. ReAct \cite{react} proposes a reasoning-then-action strategy to improve action decision accuracy. Reflexion \cite{reflexion} enhances language model agents through feedback, allowing them to reflect on task outcomes and make better decisions. However, these methods typically perform basic reasoning and do not fully exploit the capabilities of LLMs. Works like DeepSeek-R1 \cite{deepseekr1, s1} demonstrate that more comprehensive reasoning processes can significantly boost LLM performance in complex tasks.

Other studies explore replanning mechanisms to enhance task success rates. CAPE \cite{cape} uses replanning to analyze failed actions and guide the LLM to regenerate affected subplans. Text2Reaction \cite{text2reaction} adjusts local plans by comparing expected and actual feedback. These approaches focus on enabling LLMs to modify plans based on predefined error-handling strategies after failures. However, retrospective replanning is constrained by these predefined strategies and cannot inherently improve the initial task planning performance of LLMs. In contrast, LightPlanner employs hierarchical deep reasoning and dynamically parameterized skill function calls to maximize the inference capabilities of lightweight LLMs, thereby enhancing planning performance in complex scenarios.

\begin{figure*}[ht] 
\vskip 0.2in
\begin{center}
\centerline{\includegraphics[width=2\columnwidth]{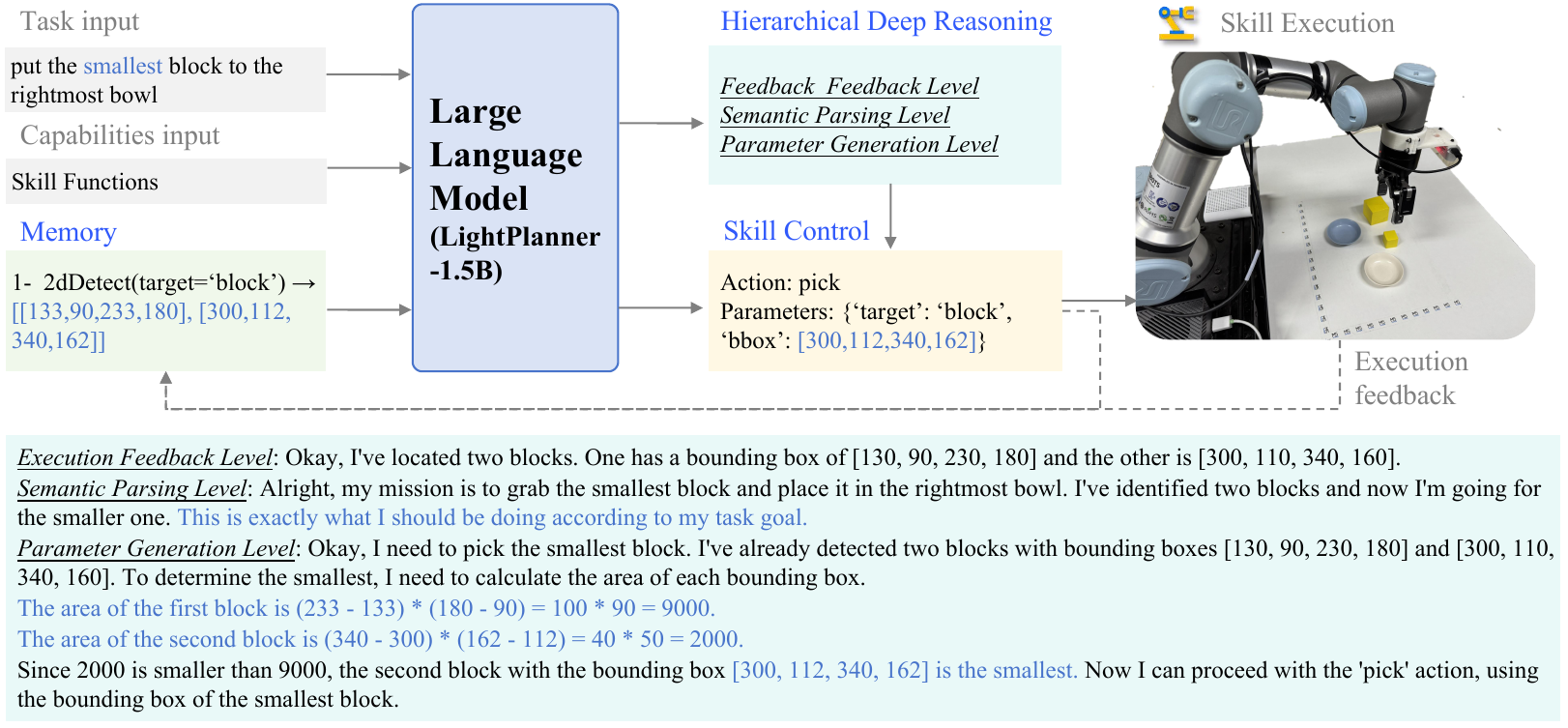}}
\caption{Architecture of LightPlanner: Generate Hierarchical Deep Reasoning and Dynamic Skill Control.}
\label{image2}
\end{center}
\vskip -0.2in
\end{figure*}

\section{Methodology}
In this section, we introduce LightPlanner, our innovative task planning framework tailored to enhance lightweight LLMs in executing complex robotic tasks. We provide a detailed description of LightPlanner's architecture and its key components, including dynamic parameterized skill control, hierarchical deep reasoning, the memory module, as well as dataset construction and model training. Each component is thoughtfully integrated to ensure efficient and robust task planning.

\subsection{Overview of LightPlanner}
Current mainstream task planners rely on large-scale, high-performance LLMs, which restrict their deployment on edge devices. Developing sophisticated task planning solutions based on lightweight LLMs remains a significant challenge in the robotics domain. LightPlanner addresses these limitations by harnessing the reasoning capabilities of lightweight LLMs to dynamically generate parameters for predefined skill functions and perform hierarchical deep reasoning to improve the correctness of action decisions. As illustrated in Figure \ref{image2}, LightPlanner integrates the task input, available skill functions, and historical execution records (memory) into the LLM. The LLM then conducts hierarchical deep reasoning and generates parameterized function calls for subsequent skills. This cohesive framework ensures that LightPlanner can effectively interpret and execute complex tasks with high accuracy and robustness, even on resource-constrained edge devices.

\subsection{Dynamic Parameterized Skill Control}
Existing LLM-based task planning methods typically rely on predefined static skill templates, which limit the system's ability to handle instructions with dynamic logical semantics. For example, commands such as “grasp the leftmost block” or “grasp the largest block” require flexible parameter generation that static templates cannot accommodate. To overcome this limitation, LightPlanner introduces a dynamic parameterized skill control strategy that enables the LLM to dynamically select and generate precise parameters for skill functions based on task requirements.

Our dynamic skill control strategy is based on three core principles:

\textbf{Separation of Perception and Action}. Traditional methods often merge perception and action into a single skill module, complicating the handling of dynamic semantics. LightPlanner decouples perception from action by treating perception algorithms as independent atomic skills that return comprehensive detection results. These results are then utilized by action skills, allowing for more flexible and accurate task execution.

\textbf{Refined Parameter Generation}. Skill functions are designed with enriched parameter inputs to facilitate precise control. For instance, as shown in Figure \ref{image2}, the \texttt{pick} skill requires not only the target object but also a bounding box (\texttt{bbox}) parameter to accurately guide the grasping position. This design enables the LLM to infer appropriate parameter values dynamically, ensuring that actions align with high-level semantic goals.
    
\textbf{Comprehensive Feedback Mechanism}. Each skill function provides detailed feedback upon execution. Perception skills return complete detection results or \texttt{None} values, while action skills return success or failure statuses along with optional fine-grained feedback from a visual language model (VLM) or a human evaluator. This feedback can enhance the hierarchical deep reasoning mechanism, enabling timely adjustments and refinements during task execution if utilized.

For example, consider the task “grasp the largest block.” LightPlanner first invokes the \texttt{2dDetect} skill to identify all blocks in the environment and obtain their bounding boxes. The LLM then processes these results to calculate the area of each bounding box, identifying the largest one. The corresponding bounding box is subsequently used as a parameter for the \texttt{pick} skill, ensuring that the grasping action targets the correct block. This dynamic parameter generation allows LightPlanner to interpret and execute complex instructions accurately.

We present pseudocode for a perception skill (\texttt{2dDetect}) and an action skill (\texttt{pick}) to illustrate the input parameters and function returns. The implementation code for all skill functions is available via the link in the supplementary materials. Here, \texttt{OWLv2} \cite{ovd} refers to an open-vocabulary detection model, and \texttt{Evaluator} is a discriminator from either a VLM or a human, providing more refined feedback. Since this paper does not focus on robotic motion planning, we employ the simplest computational methods for the robot’s grasp poses and movements. The \texttt{GetPickPos} function calculates the center point (XYZ coordinates) of the object within the \texttt{bbox} based on a depth map, while the \texttt{MoveArmTo} function provides a linear interpolation motion for the robotic arm.

\begin{algorithm}[]
\caption{2dDetect}
\begin{algorithmic}[1]
\Require Detection target \texttt{target}.
\Ensure A \texttt{bbox\_list} and feedback information.
\State \textbf{try:}
    \State \quad \texttt{bbox\_list} $\gets$ \textsc{OWLv2}(\texttt{target})
    \If{\texttt{bbox\_list} is empty}
        \State \quad \Return \{status: ``failed'', result: \texttt{None}, feedback: ``Detection result is empty.''\}
    \Else
        \State \quad \Return \{status: ``success'', result: \texttt{bbox\_list}, feedback: \texttt{None}\}
    \EndIf
\State \textbf{catch Exception e:}
    \State \quad \Return \{status: ``failed'', result: \texttt{None}, feedback: \texttt{e}\}
\end{algorithmic}
\end{algorithm}

\begin{algorithm}[]
\caption{pick}
\begin{algorithmic}[1]
\Require A target object \texttt{target}, a bounding box \texttt{bbox}.
\Ensure A pick action result, including feedback.
\State \textbf{Initialize:} \texttt{feedback\_info} $\gets$ ``''
\State \textbf{try:}
    \State \quad \texttt{pick\_pos} $\gets$ \textsc{GetPickPos}(\texttt{bbox}, \texttt{target})
    \State \quad \texttt{success\_move} $\gets$ \textsc{MoveArmTo}(\texttt{pick\_pos})
    \If{\texttt{success\_move} is false}
        \State \quad \texttt{feedback\_info} $\gets$ \textsc{Evaluator}()
        \State \quad \Return \{status: ``failed'', result: \texttt{None}, feedback: \texttt{feedback\_info}\}
    \Else
        \State \quad \Return \{status: ``success'', result: \texttt{None}, feedback: \texttt{feedback\_info}\}
    \EndIf
\State \textbf{catch Exception e:}
    \State \quad \Return \{status: ``failed'', result: \texttt{None}, feedback: \texttt{e}\}
\end{algorithmic}
\end{algorithm}

\subsection{Hierarchical Deep Reasoning}
Ensuring task success in complex scenarios necessitates robust and nuanced reasoning capabilities. LightPlanner introduces a hierarchical deep reasoning mechanism that facilitates multi-layered cognitive processes, thereby enhancing the accuracy of subsequent action decisions. This mechanism encompasses various aspects of reasoning, including execution feedback analysis, semantic understanding, and parameter validation. Error correction is integrated as a fundamental component within this broader reasoning framework, enabling the system to refine its actions based on feedback.

As illustrated in Figure \ref{image2}, the hierarchical deep reasoning mechanism operates at three distinct levels:

\textbf{Execution Feedback Level}. This level involves analyzing feedback from executed actions to inform future decisions. When an action is performed, such as a detection attempt, LightPlanner examines the feedback to refine subsequent actions. For instance, if the detection skill returns empty results, the system generates a re-calling of the detection skill as the next action step to obtain the necessary perception data. This dynamic adjustment based on skill feedback ensures that the system can adapt to unforeseen changes and maintain task progression.

\textbf{Semantic Parsing Level}. At this level, LightPlanner ensures that high-level task instructions are accurately interpreted and decomposed into actionable subgoals. For example, for the instruction “grasp the largest block,” the system verifies that the term “largest” is correctly understood within the context of the environment. This involves parsing the instruction to identify relevant attributes and ensuring that the resulting subgoals are logically consistent and aligned with the overall task objective.

\textbf{Parameter Validation Level}. Building upon semantic understanding, this level focuses on validating the parameters generated for skill functions. Taking the “grasp the largest block” example, after identifying the largest bounding box, LightPlanner verifies that the selected bounding box indeed corresponds to the largest block by recalculating areas based on perception results. This validation ensures that the parameters meet all necessary constraints and are suitable for successful action execution.

\subsection{Memory Module}

To mitigate the inference overhead caused by the continuous accumulation of context in long-term task planning with large language models (LLMs), we have integrated a memory module to record the history of task steps and their execution outcomes. Specifically, as depicted in Figure \ref{image2}, the memory module tracks the skill controls and execution results at each step. During each inference, the historical data stored in the memory module guides the current reasoning process, ensuring that the system maintains an awareness of past actions and their outcomes without succumbing to the inefficiencies associated with extensive context accumulation.

\subsection{LightPlan-40k Dataset and Model Training}

\textbf{LightPlan-40k Dataset Construction}. We constructed the LightPlan-40k dataset in three stages. First, we manually annotated 60 tasks, including 30 standard tasks without logical semantic reasoning and 30 tasks with logical semantic reasoning, based on 13 predefined skill functions. Each task comprised action chains of 2 to 13 steps, with experts recording the reasoning and skill function calls and manually executing the tasks on a real robot to ensure accuracy.

In the second stage, we expanded the dataset to 6,000 tasks using GPT-4 \cite{gpt4}. This augmentation involved replacing task objects, modifying perception results, and adjusting action function parameters through predefined prompts.

Finally, we transformed the augmented tasks by isolating each action control step as an individual data instance, resulting in a comprehensive training set of 40,000 decision samples (Figure \ref{image2} shows an example of a one-step action decision). All training data are open-sourced in the supplementary materials.

\textbf{Model Training}. We fine-tuned the Qwen2.5-1.5B-Instruct \cite{qwen2} large language model using the LightPlan-40k dataset, combined with the LIMA \cite{lima} and OpenR1-Math-94k \cite{openr1} datasets. The training employed a learning rate of $4 \times 10^{-5}$ over three epochs with full-parameter tuning. The resulting model is named LightPlanner-1.5B.

\textbf{Supplementary Materials}. All training data are available in the supplementary materials.

\begin{table*}[ht]
\caption{Average Success Rates (SR) and Task Completion Rates (CR) for Real-World Manipulation Tasks, Categorized by Task Length. Bold indicates the best performance, and underlined indicates the second best.}
\label{table2}
\vskip 0.15in
\begin{center}
\begin{small}
\begin{sc}
\begin{tabular}{lllcccccc}
\toprule
& & &  \multicolumn{2}{c}{\textbf{Short-term}} & \multicolumn{2}{c}{\textbf{Medium-term}} & \multicolumn{2}{c}{\textbf{Long-term}} \\
\midrule
\textbf{Method} &\textbf{LLM} &\textbf{Size} & \textbf{\%SR$\uparrow$} & \textbf{\%CR$\uparrow$} & \textbf{\%SR$\uparrow$} & \textbf{\%CR$\uparrow$} & \textbf{\%SR$\uparrow$} & \textbf{\%CR$\uparrow$} \\
\midrule
\multicolumn{9}{l}{\textbf{Simple Semantic Mapping Tasks}} \\
\midrule
CaP                 & GPT-3  & 175B & 97.2   & 98.1  & 83.2 & 84.6 & 64.1 & 68.3  \\
CAPE                & GPT-4  & -    & 100.0  & 100.0 & 83.5 & \underline{87.0} & 66.7 & 72.6  \\
CoPAL               & GPT-4  & -    & \textbf{100.0}  & \textbf{100.0} & \textbf{85.4} & \textbf{89.3} & \textbf{71.2} & \underline{73.3}  \\
MLDT                & LLaMA2 & 7B   & 100.0  & 100.0 & 77.1 & 79.3 & 66.4 & 69.5  \\
ReAct-1.5B          & Qwen2.5& 1.5B & 100.0  & 100.0 & 77.4 & 80.1 & 65.0 & 68.2  \\
ReAct-3B            & Qwen2.5& 3B   & 100.0  & 100.0 & 80.3 & 82.0 & 67.3 & 71.0  \\
\textbf{LightPlanner-1.5B} & Qwen2.5& 1.5B & \textbf{100.0}  & \textbf{100.0} & \underline{83.6} & \textbf{89.3} & \underline{70.5} & \textbf{73.6}  \\
\midrule
\multicolumn{9}{l}{\textbf{Dynamic Semantic Reasoning Tasks}} \\
\midrule
CaP                 & GPT-3  & 175B & 83.1  & 84.6 & 66.2 & 68.6 & 48.1 & 53.2  \\
CAPE                & GPT-4  & -    & 85.2  & 87.1 & 68.2 & 69.9 & \underline{54.1} & 56.6  \\
CoPAL               & GPT-4  & -    & \underline{86.1}  & \underline{87.3} & \underline{71.4} & \underline{73.1} & 54.0 & \underline{56.8}  \\
MLDT                & LLaMA2 & 7B   & 81.4  & 83.9 & 65.1 & 66.0 & 47.4 & 49.4  \\
ReAct-1.5B          & Qwen2.5& 1.5B & 80.0  & 82.7 & 64.3 & 67.2 & 48.3 & 49.2  \\
ReAct-3B            & Qwen2.5& 3B   & 82.1  & 83.4 & 66.5 & 68.1 & 51.2 & 53.8  \\
\textbf{LightPlanner-1.5B} & Qwen2.5& 1.5B & \textbf{100.0}  & \textbf{100.0} & \textbf{79.5} & \textbf{83.1} & \textbf{66.1} & \textbf{68.3}  \\
\cellcolor{gray!25}GPT-4 + LightPlanner & GPT-4  & -   & 100.0  & 100.0 & 85.1 & 89.2 & 75.8 & 80.0  \\
\bottomrule
\end{tabular}
\end{sc}
\end{small}
\end{center}
\vskip -0.1in
\end{table*}

\section{EXPERIMENTS}

\subsection{Experimental Setup}

\begin{figure*}[ht] 
\vskip 0.2in
\begin{center}
\centerline{\includegraphics[width=2\columnwidth]{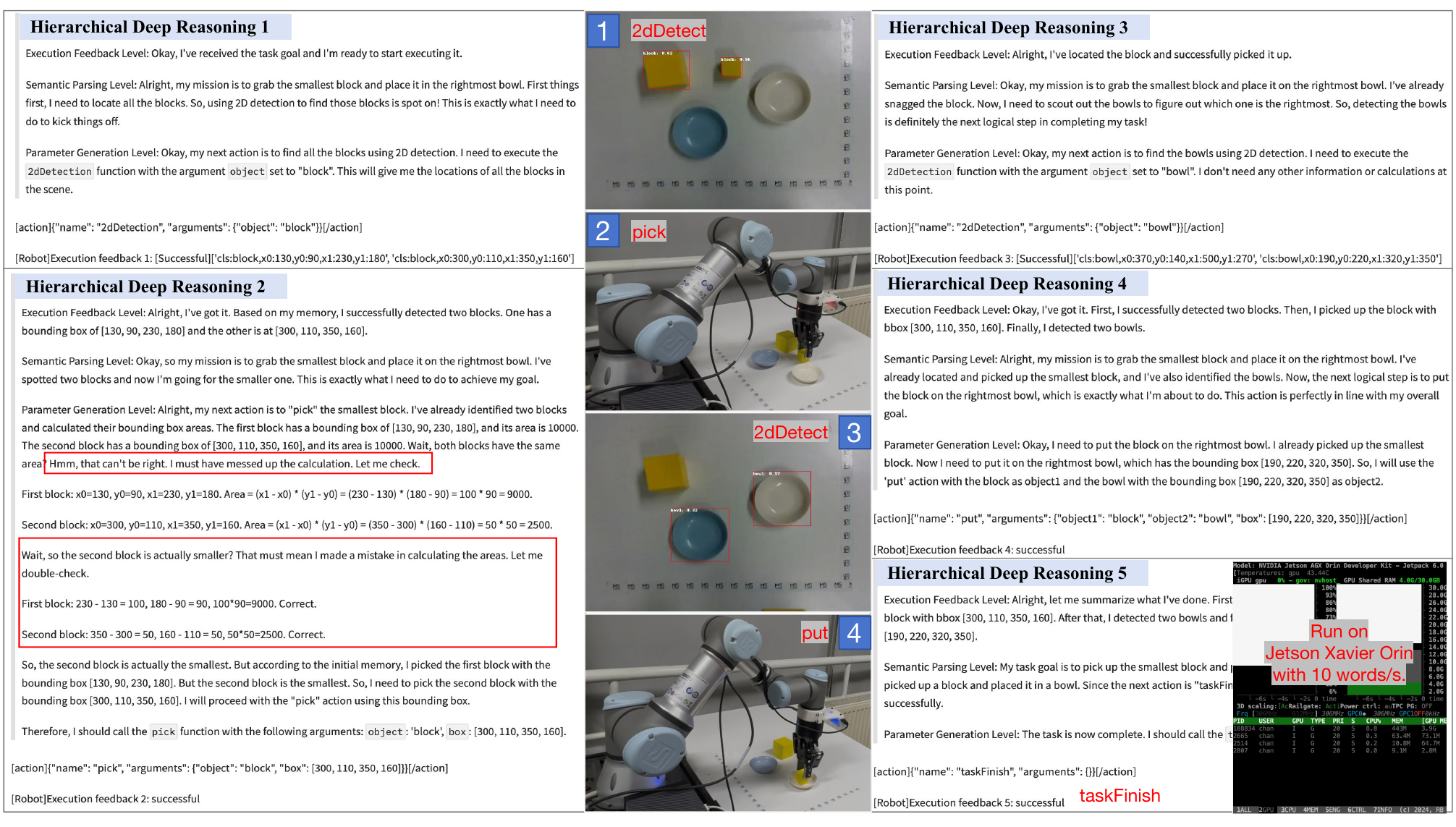}}
\caption{A Complete Example. LightPlanner's efficient performance on Jetson Xavier Orin.}
\label{image6}
\end{center}
\vskip -0.2in
\end{figure*}

\textbf{Hardware and Environment.} As depicted in Figure \ref{image2} and \ref{image6}, we established a real-world environment using a UR3 robot equipped with a 6-DoF arm and a 1-DoF Robotiq gripper. An Intel D435 RGB-D camera is mounted on the robot arm's wrist. For quantitative experiments, all software components, including the LightPlanner model, are executed locally on an RTX 3090 GPU.

\textbf{Evaluation Tasks.} We assessed LightPlanner on 30 distinct tasks, each repeated three times under varying lighting and background conditions, resulting in a total of 90 task trials. The tasks were categorized into two subsets: (1) \textbf{Simple Semantic Mapping Tasks}, consisting of 15 tasks with clearly defined semantic goals in the environment, such as picking up a single block when only one is present, requiring minimal logical reasoning; and (2) \textbf{Dynamic Semantic Reasoning Tasks}, consisting of 15 tasks requiring dynamic semantic reasoning, such as interpreting the word “largest” in “pick up the largest block” or distinguishing spatially specific objects (e.g., “pick up the leftmost block”). The complexity of the tasks varies, with action chain lengths ranging from 2 to 8 steps. Each subset includes 5 short-term tasks (2-4 steps), 5 medium-term tasks (5-6 steps), and 5 long-term tasks (7-8 steps).

\textbf{Evaluation Metrics.} 
\begin{itemize}
    \item \textit{Task Success Rate (SR)}: The percentage of tasks successfully completed without any failures.
    \item \textit{Average Task Completion Rate (CR)}: The ratio of the length of the action chain completed before a failure occurs to the total length of the task.
\end{itemize}

\textbf{Baselines.} 
\begin{itemize}
    \item \textit{CaP (Code as Policies)} \cite{cap}: A task planner leveraging ChatGPT to generate executable code that invokes predefined skill APIs based on task instructions.
    \item \textit{CAPE} \cite{cape}: A GPT-4-based task planner incorporating a replanning strategy.
    \item \textit{CoPAL} \cite{copal}: A task planning system utilizing GPT-4 to generate multi-level task plans through agent-based methods.
    \item \textit{MLDT} \cite{plan-mldt}: A multi-level dynamic task planner focused on complex reasoning and task decomposition. It was fine-tuned from LLaMA2-7B on LightPlan-40k to create MLDT-7B.
    \item \textit{ReAct} \cite{react}: A continuous task planner that integrates reasoning and action, closely resembling LightPlanner. To ensure a fair comparison, we fine-tuned Qwen2.5-1.5B-Instruct and Qwen2.5-3B-Instruct on LightPlan-40k to obtain ReAct-1.5B and ReAct-3B, respectively.
\end{itemize}
For baselines that do not require training, we adhered to their original prompt designs and configured the associated scenes, task instructions, and skill libraries accordingly for the evaluation tasks. For ReAct and MLDT, we converted LightPlan-40k into the required data formats and subsequently trained them using the original parameter settings.

\subsection{Experimental Results and Analysis}

The experimental results presented in Table \ref{table2} demonstrate the outstanding performance of LightPlanner across various tasks. In simple semantic mapping tasks, LightPlanner-1.5B achieved a 100\% success rate for short-term tasks. For medium-term tasks, it attained a success rate (SR) of 83.6\%, outperforming ReAct-3B, which achieved 80.3\%. In long-term tasks, LightPlanner-1.5B maintained a superior SR of 70.5\% compared to ReAct-3B's 67.3\%. These results indicate that our approach can sustain high planning accuracy through effective dynamic skill control, even with fewer parameters. Notably, in dynamic semantic reasoning tasks, LightPlanner-1.5B achieved a 100\% success rate in short-term tasks and significantly outperformed all baseline methods in medium-term (79.5\% SR vs. 71.4\% for CoPAL) and long-term tasks (66.1\% SR vs. 54.1\% for CAPE). From these findings, we draw two key conclusions. First, LightPlanner's dynamic parameterization accurately interprets spatial semantics, such as "largest," by calculating the active area in detection results. In contrast, template-based methods cannot handle such dynamic attributes. Second, the hierarchical reasoning mechanism prevents error propagation by verifying semantic consistency and parameter validity at each step, which is particularly crucial for long-term tasks.

To further validate the versatility and upper limits of our method, we conducted an additional set of experiments combining GPT-4 with LightPlanner (GPT-4 + LightPlanner). We prompted GPT-4 to adopt LightPlanner's approach for reasoning and planning, incorporating a context-based learning strategy with single examples from different task types. The results shown in Table \ref{table2} indicate that the success rate and completion rate in long-term tasks increased to 75.8\% and 80.0\%, respectively, marking a substantial overall performance improvement. Based on these results, we conclude that: (1) The combination of LightPlanner with GPT-4 demonstrates the universal applicability of our method; (2) Despite LightPlanner-1.5B having significantly fewer parameters than GPT-4, it achieved a 66.1\% success rate in long-term tasks, underscoring the efficiency of our approach in extracting reasoning capabilities from lightweight LLMs.

\begin{table}[]
\begin{center}
\begin{sc}
\caption{Results of Generalization Experiments on Unseen Tasks.}
\label{table3}
\begin{tabular}{lccc}
\toprule
\textbf{Module} & \textbf{LLM} & \textbf{\%SR$\uparrow$}  & \textbf{\%CR$\uparrow$}\\
\midrule
CoPAL       & GPT-4 & 53.8 & 56.8 \\
ReAct-3B & Qwen2.5 & 42.0 & 45.4  \\
LightPlanner-1.5B & Qwen2.5 & 62.6 & 64.0 \\
\bottomrule
\end{tabular}
\end{sc}
\end{center}
\vskip -0.1in
\end{table}

\textbf{Generalization Experiments.} We conducted a generalization experiment by adding 10 task categories with an action chain length of 8 that do not overlap with the training set (i.e., the action chain orders are not seen during training). We compared LightPlanner with GPT-4-based CoPAL and ReAct-3B fine-tuned on an open-source LLM. As shown in Table \ref{table3}, LightPlanner achieved the highest task success rate and task completion rate, demonstrating superior generalization capabilities.

\begin{table}[]
\begin{center}
\begin{sc}
\caption{Ablation Study Results in Long-term Tasks.}
\label{table4}
\begin{tabular}{lccc}
\toprule
\textbf{Module} & \textbf{\%SR} & \textbf{\%CR} & \textbf{Time(s)} \\
\midrule
Full LightPlanner & 66.1 & 68.3 & 36 \\
w/o Dynamic Skill Control & 58.6 & 59.2 & 36 \\
w/o Hierarchical Deep Reasoning & 60.3 & 63.6 & 24 \\
w/o Memory Module & 61.1 & 63.9 & 40 \\
\bottomrule
\end{tabular}
\end{sc}
\end{center}
\vskip -0.1in
\end{table}

\begin{figure}[ht] 
\vskip 0.2in
\begin{center}
\centerline{\includegraphics[width=0.9\columnwidth]{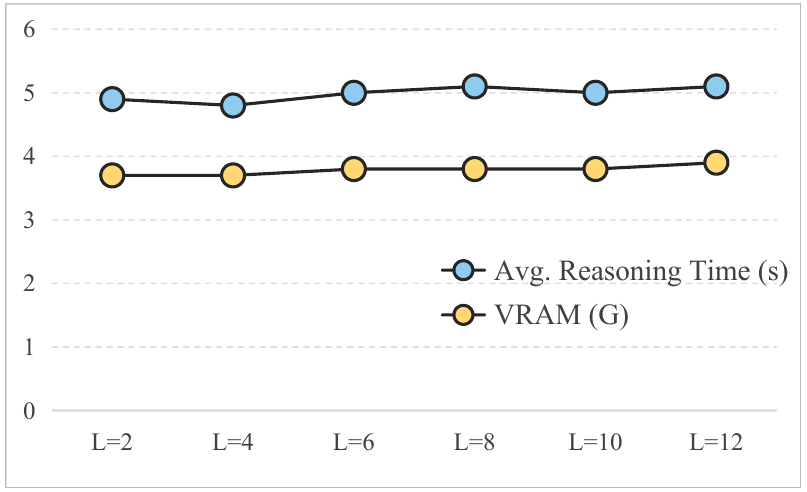}}
\caption{Average reasoning latency and VRAM consumption per round of a skill for tasks ranging from simple to complex. Here, \texttt{L} denotes the length of the task's action chain.}
\label{image5}
\end{center}
\vskip -0.2in
\end{figure}

\subsection{Ablation Study}
To validate the contributions of each component in LightPlanner, we conducted a comprehensive ablation study by removing key modules and evaluating their impact on task success rates (SR), completion rates (CR), and total execution time. The results, presented in Table \ref{table4}, highlight the critical role of each component in maintaining system performance.

\textbf{Impact of Dynamic Skill Control.} Disabling dynamic parameter generation (reverting to static templates) resulted in the poorest performance, with SR decreasing by 7.5\% and CR by 9.1\%. This indicates that our parameterized skill control significantly enhances the system's ability to handle dynamic semantics through active parameter inference, thereby overcoming the limitations of fixed templates.

\textbf{Impact of Hierarchical Deep Reasoning.} Removing this module led to a substantial decline in performance, with SR dropping by 5.8\% (from 66.1\% to 60.3\%) and CR decreasing by 4.7\%. This demonstrates that our three-level reasoning mechanism (semantic parsing, parameter validation, and execution feedback) effectively prevents error propagation by enabling early error detection and correction.

\textbf{Impact of Memory Module.} Without the memory module, SR decreased by 5.0\% and CR by 4.4\%, accompanied by a 4-second increase in total execution time. The time degradation arises from the LLM needing to process the full historical context for each planning step. This confirms that our memory mechanism successfully reduces redundant context processing while preserving essential historical information, thereby ensuring stable and efficient long-term planning.

These findings collectively demonstrate that the three core components synergistically contribute to LightPlanner's superior performance: dynamic skill control provides essential flexibility for semantic interpretation, hierarchical deep reasoning ensures action correctness through multi-level verification, and the memory module optimizes efficiency in long-term planning. The complementary effects of these components underscore their importance in achieving robust and efficient task planning.

\subsection{Analysis of Planning Efficiency and Resource Consumption}

\textbf{Memory Module Gain in Time.} In long-term tasks, existing methods often incrementally accumulate historical context information, leading to a substantial increase in input tokens and causing planning time to grow linearly with the number of task steps. In contrast, LightPlanner utilizes a memory module to iteratively update historical information, retaining only essential key data and ensuring that the input text for each inference remains minimal. As illustrated in Figure \ref{image5}, regardless of task complexity, LightPlanner’s single-round planning time varies by only $\pm$2 seconds, maintaining stability and achieving more efficient real-time inference in long-term tasks. Additionally, LightPlanner’s memory usage remains stable at approximately 3.9 GB.

\textbf{Edge Platform Deployment and Case Study.} Figure \ref{image6} demonstrates LightPlanner running efficiently on the NVIDIA Jetson Xavier Orin, achieving a processing speed of 10 words/s with a maximum memory usage of 3.9 GB. The showcased task involves grasping the rightmost block and placing it onto the largest block. Successfully completing this task requires inferring the bounding boxes (bboxes) of both blocks during the grasping and placing operations. Additionally, below Figure \ref{image6}, LightPlanner’s multi-level error correction reasoning process is illustrated. This edge deployment example underscores that our fine-tuned LLM effectively handles scenarios demanding high real-time performance and local deployment, thereby offering significant application value.

\section{Conclusion}
In this paper, we present LightPlanner, a novel task planner designed to enhance the performance of lightweight large language models (LLMs) in complex robotic task planning. By integrating dynamic parameterized skill control, hierarchical deep reasoning, and a memory module, LightPlanner significantly improves task success rates while maintaining high efficiency, even on resource-constrained edge devices. Our experiments demonstrate that LightPlanner outperforms existing methods in both simple and dynamic semantic reasoning tasks, showcasing its robustness and generalization capabilities. While LightPlanner achieves remarkable performance in static environments, it may encounter challenges in highly dynamic scenarios where objects or environments change unpredictably during task execution. To further enhance the reasoning capabilities of lightweight LLMs, we plan to explore reinforcement learning–based training methods inspired by approaches like DeepSeek-R1. By incorporating reinforcement learning, we aim to improve the model's adaptability and decision-making skills in complex and dynamic environments. Additionally, we will investigate the potential of multi-agent task planning to enable more sophisticated collaborative robotics applications.

\normalem

\bibliographystyle{ieeeconf/IEEEtran}
\bibliography{ieeeconf/IEEEabrv,ieeeconf/main}

\end{document}